\documentclass{article}
\usepackage{mkstyle}
\usepackage{latexsym}

\newcommand{\newstaff}[1]{#1}
\newcommand{\Bem}[1]{}
\Bem{
}
\usepackage{graphicx}
\usepackage{emcsr96}

\title{
Query Optimization Properties
 of Modified Valuation-Based Systems 
}
\shorttitle{
Query Optimization Properties of Modified VBS
}
\author{Mieczys{\l}aw A. K{\l}opotek\\
Institute of Computer Science\\
Polish Academy of Sciences\\
ul. Ordona 21, \\
01-237 Warszawa, Poland\\
{\small email: klopotek{@}ipipan.waw.pl} \\
\And S{\l}awomir T. Wierzcho\'{n}\\
Institute of Computer Science\\
Polish Academy of Sciences\\
ul. Ordona 21, \\
01-237 Warszawa, Poland\\
{\small email: stw{@}ipipan.waw.pl} \\
}

      \newtheorem{lemma}[thm]{Lemma}

\newcommand{\rwk}{\mbox{$\bigcirc\hspace*{-1.9ex}\mbox{\scriptsize\rm R}\,$}}
\newcommand{\remove}{\rwk}
\newcommand{\uu}[1]{\mbox{{$\cal #1$}}}

\newcommand{\ax}[1]{{\bf #1}}

\newcommand{\bfS}{\mbox{\bf S}}
\newcommand{\bfH}{\mbox{\bf H}}
\newcommand{\bfE}{\mbox{\bf E}}
\newcommand{\bfN}{\mbox{\bf N}}
\newcommand{\bfF}{\mbox{\bf F}}

\date{}

\begin{document}
\maketitle

\begin{abstract}
Valuation-Based~System  can
represent \linebreak
 knowledge in different domains including probability
theory, Dempster-Shafer theory and possibility theory. More
recent studies show that the framework of VBS is also
appropriate for representing and solving Bayesian decision
problems and optimization problems. 
  In this
paper after introducing the valuation based system (VBS) framework, 
 we present Markov-like properties of VBS 
and a method for resolving queries to VBS. 

\Bem{

{\bf Symposium:}  D Fuzzy systems, Approximate Reasoning and Knowledge-based
Systems. 
}
\end{abstract}

\section{
 Introduction
}

 Though graphical representation of a domain knowledge has
quite long history, its full potential has not been recognized
until
recently. We should mention here pioneering works of J.~Pearl,
reported in his monography published in 1988 \shortcite{Pearl:88}. Further
development in this domain has been achieved by Shenoy and
Shafer \shortcite{Shenoy:86} who adopted a method used in solving nonserial
dynamic programming problems 
\cite{Bertele:72}. This
trick proved   to be very fruitful and gave growth to a unified
framework for uncertainty representation and reasoning, called
{\bf V}aluation-{\bf B}ased {\bf S}ystem,
 VBS for short \cite{Shenoy:89}. It can
represent knowledge in different domains including probability
theory, Dempster-Shafer theory and possibility theory. More
recent studies show that the framework of VBS is also
appropriate for representing and solving Bayesian decision
problems \cite{Shenoy:93} and optimization problems 
\cite{Shenoy:91}. The graphical representation is called a 
{\it valuation
network}, and the method for solving problems is called the
{\it fusion algorithm}.  Closely related to VBS is the algorithm of
Lauritzen and Spiegelhalter \shortcite{Lauritzen:88} and HUGIN approach
developed by Jensen and his  co-workers \shortcite{Jensen:90}. 

 A Bayesian network (as well as its generalization - 
VBS)  can be
regarded as a summary of an expert's experience with an implicit
population. Detailed documentation of such knowledge with an
explicit population is stored in a database.  It appears that
there exists a strong connection between these two approaches.
First of all, databases are used for knowledge acquisition and
Bayesian network identification - see \cite{Pearl:88} or 
\cite{Cooper:92} for a deeper discussion. Studies by Wen
\shortcite{Wen:91}, and Wong, Xiang and Nie \shortcite{Wong:93} establish a
link between knowledge-based systems for probabilistic reasoning and
relational databases. Particularly, they show that the belief
update in a Bayesian network can be processed as an ordinary
query, and the techniques for query optimization are directly
applicable to updating beliefs. The same idea we find in Thoma's
\shortcite{Thoma:91} works, who proposed a scheme for storing Shafer's
belief functions which generalizes graphical models.  

 In this
paper after introducing the valuation based system framework
(Section 2),  we present Markov-like properties of VBS (Section
3) and a method for resolving queries to VBS (Section 4).

\section{
 Valuation Based Systems
}

 The VBS framework was introduced in \cite{Shenoy:89}. In VBS, a
domain knowledge is represented by entities called {\it variables } and
{\it valuations}. Further, two operations called {\it combination } and 
{\it marginalization }  are defined on valuations to perform a local
computational method for computing marginals of the joint
valuation.  The basic components of VBS can be characterized as
follows.

\quad\\
{\bf 
Valuations
}\\

 Let  $\uu{X}=\{x_1,x_2,...x_n\}$ be a finite set of variables and $\Theta_i$
be the domain (called also {\it frame}), i.e. a discrete set of possible
values of i-th variable. If h is a finite non-empty set of 
variables then $\Theta(h)$ denotes the Cartesian product of $\Theta_i$  for
$x_i$ in $h$, i.e. $\Theta(h) = \times \{\Theta_i|x_i \in h\}$.
 \uu{R} stands for
a set of non-negative reals. For each subset s of \uu{X} there is a set $D(s)$
called the domain of a valuation. For instance in the case of
probabilistic systems $D(s)$ equals to $\Theta(s)$, while under the
belief function framework $D(s)$ equals to the power set of  $\Theta(s)$,
i.e. $D(s) = 2^{\Theta(s)}$. Valuations, being primitives in the VBS
framework, can be characterized as mappings $\sigma: D(s) \rightarrow
\uu{R}$. In
the sequel valuations will be denoted by lower-case Greek letters,
$\rho$, $\sigma$, $\tau$, and so on. Following Shenoy 
\shortcite{Shenoy:94} we distinguish
three categories of valuations:

\begin{itemize}
\item 
 {\it Proper valuations}, \uu{P}, represent knowledge that is partially
coherent. (Coherent knowledge means knowledge that has well
defined semantics.) This notion plays an important role in the
theory of belief functions: by proper valuation it is understood
an unnormalized commonality function.
\item 
 {\it Normal valuations}, \uu{N}, represent another kind of partially
coherent knowledge. For instance, in probability theory, a
normal valuation is a function whose values sum to 1.
Particularly, the elements of $\uu{P} \cap \uu{N}$ are called  proper normal
valuations; they represent knowledge that is completely coherent
or knowledge that has well-defined semantics. 
\item 
 {\it Positive normal valuations}: it is a subset $\uu{U}_s$ of $\uu{N}_s$
consisting of  all valuations that have unique identities in  $\uu{N}_s$.

\end{itemize}

Further there are two types of special valuations:

\begin{itemize}
\item 
 {\it Zero valuations} represent knowledge that is internally
inconsistent, i.e. knowledge whose truth value is always false;
e.g., in probability theory by zero valuation we understand a
valuation that is identically zero. It is assumed that for each
$s \subseteq \uu{X}$ there is at most one valuation $\zeta_s \in 
\uu{V}_s$ . The
set of all zero valuations is denoted by~\uu{Z}.
\item 
 {\it Identity valuations}, I, represent total ignorance, i.e. lack
of knowledge. In probability theory an identity valuation
corresponds to the uniform probability distribution. It is
assumed that for each $s \subseteq \uu{X}$ the commutative semigroup (w.r.t.
the binary operation $\otimes$  defined later) $\uu{N}_s\cup \{\zeta_s\}$ has
an identity $\iota_s \in \uu{V}_s$ . Commutative semigroup may have at most
one identity \cite{Clifford:61}.
\end{itemize}

\quad\\
{\bf Combination
}\\

 By combination we understand a mapping $\otimes :\uu{V} \times \uu{V} 
 \rightarrow \uu{N} \cup  \uu{Z}$ that
satisfies the following six axioms:

\begin{description}
\item[\ax{(C1)}] If $\rho \in \uu{V}_r$ and $\sigma \in \uu{V}_s$ then 
$\rho \otimes  \sigma \in \uu{V}_{r\cup s}$;

\item[\ax{(C2)}] $\rho \otimes  (\sigma \otimes \tau) = 
(\rho \otimes  \sigma) \otimes  \tau$;
 
\item[\ax{(C3)}] $\rho \otimes  \sigma = \sigma \otimes  \rho$; 

\item[\ax{(C4)}] If  $\rho \in \uu{V}_r$ and zero valuation $\zeta_s$ 
exists then $\rho \otimes  \zeta_s \in \uu{V}_{r\cup s}$.

\item[\ax{(C5)}] For each $s \subseteq \uu{X}$ there 
exists an identity valuation $\iota_s \in  \uu{N}_s
\cup \{\zeta_s\}$ such that for each valuation 
$\sigma \in  \uu{N}_s \cup \{\zeta_s\}$, $\sigma \otimes  \iota_s = \sigma$.

\item[\ax{(C6)}] It is assumed that the set $\uu{N}_\emptyset$  consists of
exactly one element denoted $\iota_\emptyset$ .
\end{description}

 In practice combination of two valuations is implemented 
as follows. Let (+) be a binary operation on \uu{R}. Then 
$(\sigma \otimes  \rho)(x) =
\sigma(x.s) (+) \rho(x.r)$ where $x$ is an element from $D(s)$ and $x.r$, 
$x.s$
stand for the projection (relying upon dropping unnecessary
variables) of $x$ onto the appropriate domain $D(r)$ or $D(s)$. In
probability theory combination corresponds to pointwise
multiplication followed by normalization, and in Dempster-Shafer
theory to the Dempster rule of combination.

 In the field of uncertain reasoning combination corresponds to
aggregation of knowledge: when $\rho$ and $\sigma$ represent our knowledge
about variables in subsets $r$ and $s$ of  
\uu{X} then the valuation $\rho \otimes 
\sigma$ represents the aggregated knowledge about variables in $r  \cup  s$.
Moreover Wen \shortcite{Wen:91}, and Wong, Xiang and Nie 
\shortcite{Wong:93} showed that
under probabilistic context combination corresponds to the
(generalized) join operation used in the data-based systems.
Hence the belief update in a Bayesian network can be processed
as an ordinary query, and the techniques for query optimization
are directly applicable to updating beliefs. Similar idea we
find in Thoma's \shortcite{Thoma:91} works, who proposed a scheme for
storing Shafer's belief functions. 

 If $\rho \otimes  \sigma$ is a zero valuation, we say that $\rho$ and
$\sigma$  are
inconsistent. On the other hand, if $\rho \otimes  \sigma$ is a normal
valuation, then we say that $\rho$ and $\sigma$ are consistent.

 It is important to notice, that an implication of axioms \ax{C1 -
C3} is that the set $\uu{N}_s \cup \{\zeta_s\}$ together with the combination
operator is a commutative semigroup 
 \cite{Clifford:61}.
 If zero valuation $\zeta_s$ exists then $\zeta_s$ is - by axiom \ax{C4} -
the zero of this semigroup. Similarly, by axiom \ax{C5}, the identity
valuation is the identity of the semigroup $\uu{N}_s \cup \{\zeta_s\}$.

\quad\\
{\bf Marginalization}\\

 While combination results in knowledge expansion,
marginalization results in knowledge contraction. Let $s$ be a
non-empty subset of  \uu{X}. It is assumed that for each variable X
in $s$ there is a mapping $\downarrow (s - \{X\}):\uu{V}_s \rightarrow
\uu{V}_{s-\{X\}}$,
called
marginalization to $s - \{X\}$ or deletion of $X$, that satisfies the
next six axioms:

\begin{description}

\item[\ax{(M1)}] Suppose $\sigma \in \uu{V}_s$ and suppose $X, Y \in  s$. 
 Then\\
 $(\sigma ^{\downarrow (s - \{X\})})^{\downarrow (s
- \{X,Y\})} = (\sigma ^{\downarrow (s - \{Y\})}
)^{\downarrow (s - \{X,Y\})}$ ;

\item[\ax{(M2)}] If zero valuation exists, then 
$\zeta_s ^{\downarrow (s - \{X\})} = \zeta_{s-\{X\}}$;

\item[\ax{(M3)}] 
$\sigma ^{\downarrow (s - {X})} \in  \uu{N}$ if and only if 
$\sigma \in  \uu{N}$ ;

\item[\ax{(M4)}] If $\sigma \in  \uu{U}$ 
then $\sigma ^{\downarrow (s - {X})} \in  \uu{U}$;

\item[\ax{(CM1)}] Suppose $\rho \in \uu{V}_r$ and $\sigma \in \uu{V}_s$. 
Suppose $X \not\in  r$ and $X \in  s$. Then

 $(\rho \otimes  \sigma)^{\downarrow ((r \cup  s) - \{X\})} = 
\rho \otimes
\sigma ^{\downarrow ( s - \{X\})}$ 

\item[\ax{(CM2)}] Suppose $\sigma \in  \uu{N}_s$. Suppose 
$r \subseteq s$ and suppose that $\iota$  is an
identity for $\sigma ^{\downarrow r}$. Then

 $\sigma \otimes  \iota = \sigma$. 

\end{description}

 Axiom \ax{M1}   states that if we delete from s, the domain of a
valuation $s \in \uu{V}_s$, two variables, say $X$ and $Y$, then the resulting
valuation defined over the subset $r = s - \{X,Y\}$ is invariant to
the order of these variables deletion. Particularly, deleting
all variables from the set s we obtain the valuation whose
domain is the empty set (its existence is guaranteed by axiom
\ax{C6}); by axiom \ax{M3} this element equals to $\iota_\emptyset$  if and
only if $\sigma$ is a normal valuation. 

 Axioms \ax{M2 - M4} state that the marginalization preserves
coherence of knowledge. Axiom \ax{CM1} plays an important role in
designing the Message Passing Algorithm (MPA, for short) which
will be described later, and axiom \ax{CM2} allows to characterize
properties of the identity valuations; some of them are given in
the Lemma \ref{l21} below.

\begin{lemma} \label{l21} 
  \cite{Shenoy:94}. If axioms \ax{C1 - C6, M1 - M4, CM1} and
\ax{CM2} are satisfied then the following statements hold.

 1. Let $\sigma \in \uu{V}_s$ and $r \subseteq s$. $\sigma \in  \uu{N}_s \cup
\{\zeta_s\}$ if and only if $\sigma \otimes  \iota_r = \sigma$.

 2. If $\sigma \in \uu{V}_s$ and $r  \subseteq  s$  then  $\sigma 
\otimes  \iota_r = 
\sigma \otimes
\iota_\emptyset$ . 

 3. $\iota_s \otimes  \iota_r = \iota_{s \cup  r}$.

 4. If  $r \subseteq s$ then $\iota_s ^{\downarrow r} = \iota_r$. 
\end{lemma}

\quad\\
{\bf Removal
}\\

\begin{figure}
\begin{center}
\Bem{
\begin{picture}(12,4.3)
\put(0,4.3){\special{em:graph hgh1.pcx}}
\end{picture}
}
%
     \includegraphics[width=0.4\textwidth]{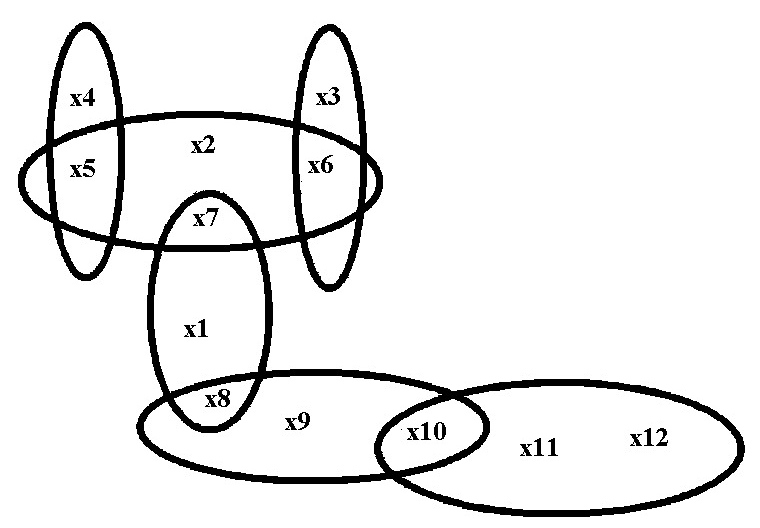}
\end{center}
\caption{Hypergraph $H_1$ - graphical representation.}
\label{hypergraph}
\end{figure}
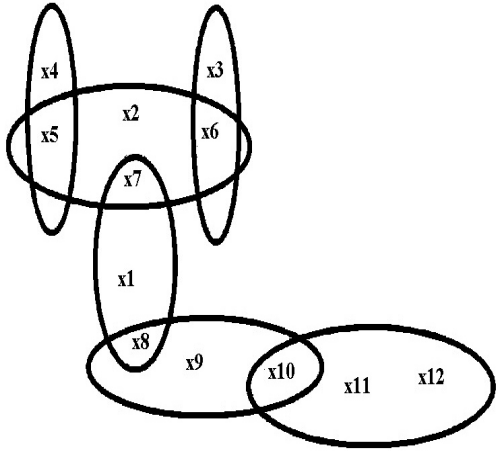

 Removal, called also direct difference, is an "inverse"
operation to the combination. Formally, it can be defined as a
mapping $\remove :\uu{V} \times  (\uu{N} \cup  \uu{Z}) 
\rightarrow \uu{N} \cup  \uu{Z}$, that satisfies the
three axioms: 

\begin{description} 
\item[\ax{(R1)}] If  $\sigma \in \uu{V}_s$ and $\rho \in  \uu{N}_r\cup  
\uu{Z}_r$ then 
$\sigma \remove\rho \in  \uu{N}_{r\cup s}\cup  \uu{Z}_{r\cup s}$.
\item[\ax{(R2)}] For each $\rho \in  \uu{N}_r\cup  \uu{Z}_r$ and for each $r
\subseteq \uu{X}$
there exists an identity $\iota_r$ such that $\rho \remove  \rho = 
\iota_r$ .
\item[\ax{(CR)}] If  $\sigma, \tau \in \uu{V}$ and $\rho \in  \uu{N} \cup  
\uu{Z}$ then 
$(\sigma \otimes  \tau)
\remove \rho =  \sigma \otimes  (\tau \remove  \rho)$.
\end{description}

 Note that we can define the (pseudo)-inverse of a normal
valuation by setting  
$\rho ^{-1} = \iota_\emptyset  \remove  \rho$. The main properties of
removal are summarized in Lemma \ref{l22} given below.

\begin{lemma} \label{l22} 
\cite{Shenoy:94}.  Suppose that $\sigma,\tau \in \uu{V} $and $\rho \in  \uu{N}
 \cup
\uu{Z}$. Then:

 1.  $(\sigma \otimes  \tau) \remove  \rho 
= (\sigma \remove  \rho\otimes  \tau) \otimes  \tau$.

 2. If 
$\sigma \in \uu{V}_s$  and $r \subseteq s$, 
then $\sigma \remove  \iota_r = \sigma \otimes
\iota_\emptyset  = \sigma$. 

 3. $[(\sigma \otimes  \rho) \remove  \rho] \otimes  \rho = 
  \sigma \otimes  \rho$.

 4. $\rho ^{-1} \otimes  \rho = \rho \otimes  \rho ^{-1}$.

 5. $\sigma \remove  \rho = \sigma \otimes  \rho ^{-1}$.
\end{lemma}

\quad\\
{\bf The propagation algorithm
}\\

 With the concepts already introduced we define a VBS as a
5-tuple $(\uu{X}, \bfS, 
(\sigma_s)_{s\in \bfS},\otimes ,\downarrow )$, where \bfS~
 is a family of
subsets of the set of variables \uu{X}. The aim of uncertain reasoning is to
find a marginal valuation

\begin{equation} \label{e1}
 \rho = (\otimes {\sigma_s| s\in \bfS})^{\downarrow r}, r \subseteq s, s
\in \bfS. \end{equation}

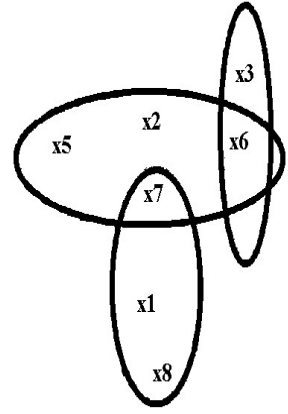
\begin{figure}
\begin{center}
\Bem{
\begin{picture}(12,4.3)
\put(0,4.3){\special{em:graph hgh2.pcx}}
\end{picture}
}
%
     \includegraphics[width=0.25\textwidth]{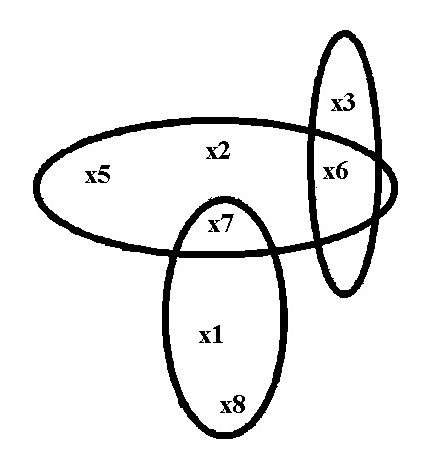}
\end{center}
\caption{Subgraph of hypergraph $H_1$ from Fig. 1
for answering query $q$.}
\label{subhypergraph}
\end{figure}

To apply the method of local computations, called the
message-passing algorithm (MPA, for brevity) observe first that
(\uu{X}, \bfS) is nothing but a hypergraph. With this hypergraph we
associate so-called Markov tree T = (\bfH,\bfE) i.e. a hypertree, or
acyclic hypergraph, (\uu{X}, \bfH), being a covering of (\uu{X}, \bfS) and
organized in a tree structure - see 
\cite{Shenoy:89} 
for details. We say that (\uu{X}, \bfH) covers (\uu{X}, \bfS) if for each 
$s$ in \bfS~
there exists $h$ in \bfH~  such
 that $s \subseteq h$. Now if  (\uu{X}, \bfH) is
a hypertree if it can be reduced to the empty set by recursively:
1) deleting vertices which are only in one edge, and 2) deleting
hyperedges which are subsets of other hyperedges. These two
steps define so-called Graham's test. The sequence of hyperedge
deletion determines a tree construction sequence (i.e. a set of
undirected edges \bfE) for a Markov tree T.
\newstaff{
Let us consider  e.g. the hypergraph $H_1=
\{\{X_1,X_7,X_8\}
, \{X_2,X_5,X_6,X_7\}
, \{X_3,X_6\} 
,$
$ \{X_4,X_5\}
,$
$ \{X_8,X_9,X_{10}\}
, \{X_{10}, X_{11}, X_{12}\}
\}$ (see Fig.\ref{hypergraph}).
We see that variables $X_1,X_3,X_4,X_{11}, X_{12}$ are contained in only one
edge. We delete them getting  
 the hypergraph 
$H_1'=
\{\{X_7,X_8\}
, \{X_2,X_5,X_6,X_7\} 
,$
$ \{X_6\}
,$
$ \{X_5\} 
,$
$ \{X_8,X_9,X_{10}\}
, \{X_{10}\}
\}$. But then hyperedges $\{X_6\}$
and  $\{X_5\}$
are contained in $\{X_2,X_5,X_6,X_7\}$, 
and  $\{X_{10}\}$ in  $\{X_8,X_9,X_{10}\}$. So we delete them getting
 $H_1''=
\{\{X_7,X_8\}
, \{X_2,X_5,X_6,X_7\}
, \{X_8,X_9,X_{10}\}
\}$. Now variables $X_2,X_5,X_6,$  $X_9,X_{10}$ are contained in only one edge
each. We get  $H_1'''=
\{\{X_7,X_8\}
,$
$ \{X_7\}
, \{X_8\}
\}$. Now  hyperedge $\{X_7,X_8\}$ contains both 
$\{X_7\}$ and $\{X_8\}$, hence we get finally  $H_1''''=
\{\{X_7,X_8\}
\}$, as the result of the Graham test which indicates that $H_1$ is a
hypertree.  
}

  Now, the message passing algorithm can be summarized as
follows: it tells the nodes of a Markov tree in what sequence to
send their messages to propagate the local information
throughout the tree. The algorithm is defined by two parts: a
fusion rule, which describes how incoming messages are combined
to make marginal valuations and outgoing messages for each node;
and a propagation algorithm, which describes how messages are
passed from node to node so that all of the local information is
globally distributed. Just as propagation takes place along the
edges of the tree, fusion takes place within the nodes. It is
important to notice that in fact the MPA coincides with the two
steps determining the tree construction sequence (i.e. Graham's
test).

\section{
 Computing marginals in a Markov tree
}

 Assume that we have constructed a Markov T = (\bfH,\bfE) tree
representative of a given VBS, and let us assign a unique number
$i \in I = \{1,2,...,n\}$, $n=$Card(\bfH), to each node in the tree. Denote
$V_i$ the original valuation stored in the $i$-th node of the tree,
and $R_j$ the resultant valuation computed for $j$-th node according
to the rule (1). Following \cite{Shenoy:89}
 this $R_j$ is
computed due to the rule\\
\begin{equation} \label{e2}
 R_j =  V_j \otimes  
(\otimes \{ M_{i\rightarrow j} | i     \in N(j)
         \}
)^{\downarrow j}
         \end{equation}\\
where $N(j)$ stands for the set of neighbours of the node $j$ in the
Markov tree, $\downarrow j$ means marginalization to the set of variables
corresponding to the node $j$, and $M_{i\rightarrow  j}$  is  the 
message sent
by node $i$ to the node j calculated according to the equation (\ref{e3})

 \begin{equation}\label{e3}
 M_{i\rightarrow j} = 
(V_i \otimes  (\otimes \{M_{k\rightarrow i}| 
                            k \in  (N(i)-\{j\} 
                                   )
                       \}
              ) 
)
^{\downarrow j        }
\end{equation}

It is obvious, that to find $R_j$ we place the node j in the root
of the Markov tree and we move successively from leaves of the
tree to its root. Note that if k is a leaf node and $\iota$ is its
neighbour, then $M_{k\rightarrow i} = (V_k)^{\downarrow i}$,
 hence (\ref{e2}) and (\ref{e3}) are defined properly.

 A disadvantage of this algorithm is such that we can compute
marginals for sets contained in the family \bfH, or for subsets of
these sets only. To find marginal for a any subset of variables
we need a more elaborated approach. This problem was studied
firstly by Xu \shortcite{Xu:95}. Below we present its more economical
modification.

 First of all we need a generalization of a set chain
representation, which has the next form under probabilistic
context  \cite{Lauritzen:88}: For a given tree
construction sequence $\{h_1,h_2,...,h_2\}$ by a separator we
understand a set $s_i$ such that $s_i  = h_i\cap(h_1\cup h_2\cup ...\cup
h_{i-1})$. Separators are easily identified in a Markov tree, namely if
$\{h_i,h_j\} \in \bfE$ then $s_i  = h_i\cap h_j$. Now, with given tree
construction sequence the joint probability distribution can be represented
as follows
\begin{equation}\label{e4}
 P(x_1,x_2,...,x_n) = R_1 \prod \{(R_i/S_i)| i = 2,...,n\}
\end{equation}
where $R_i$ and $S_i$ are the marginal probabilities defined over the
set of variables represented by the sets $h_i$ and $s_i$,
respectively. It appears, that for all  VBS's this property
can be nicely extended, as we can see below. 
First we prove a lemma on an important property of VBS removal operator
\footnote{Shenoy \shortcite{Shenoy:94} assumes implicitly this property
but does not prove it.}

\begin{lemma} \label{tm1}
In Valuation-Based Systems, the following property of
removal operator holds:
$$(\rho \remove \rho ^{\downarrow r}) \otimes \rho ^{\downarrow r} =\rho $$
\end{lemma}
\AnfBeweis

From \ax{CM2}: $\rho \otimes \iota_\emptyset=\rho$.
From \ax{CR}: $(\rho \otimes \iota_\emptyset)\remove \rho ^{\downarrow r}
      = \rho \otimes (\iota_\emptyset\remove \rho ^{\downarrow r})$. 
But by definition: $\iota_\emptyset\remove \rho ^{\downarrow r}
=(\rho ^{\downarrow r})^{-1}$, hence 
$(\rho \remove \rho ^{\downarrow r}) \otimes \rho ^{\downarrow r}=
(\rho \otimes (\rho ^{\downarrow r})^{-1}) \otimes \rho ^{\downarrow r}$. 
From \ax{C2} 
$(\rho \otimes (\rho ^{\downarrow r})^{-1}) \otimes \rho ^{\downarrow r}=
\rho \otimes ((\rho ^{\downarrow r})^{-1} \otimes \rho ^{\downarrow r})$.
But we know that: 
From \ax{R2} $\rho\remove\rho=\iota_\rho$. 
From \ax{CM2} $(\rho\otimes\iota_\emptyset)\remove\rho=\iota_\rho$. 
From \ax{CR}  $\rho\otimes(\iota_\emptyset\remove\rho)=\iota_\rho$. 
hence  $\rho\otimes\rho ^{-1}=\iota_\rho$.
Therefore 
$\rho \otimes ((\rho ^{\downarrow r})^{-1} \otimes \rho ^{\downarrow r})
=\rho \otimes \iota_{\rho ^{\downarrow r}} $.
So we get due to axiom 
\ax{CM2} $\rho \otimes \iota_{\rho ^{\downarrow r}} =\rho$.
which proves our claim. 
\EndBeweis

Now let us try to transform a Markov tree valuation
 to the form similar to 
equation (\ref{e4}). 

 Assume that we have constructed a Markov tree T = (\bfH,\bfE) 
representative of a given VBS, and let us assign a unique number
$i \in I = \{1,2,...,n\}$, $n=$Card(\bfH), to each node in the tree. Denote
$V_i$ the original valuation stored in the $i$-th node of the tree. 
Let us consider the following transformation algorithm: starting with the
node $k$=$n$ down to $1$ we run a "valuation move" step such that
we will "move" valuation from nodes with smaller number $i$ to ones
with higher one so that final  valuation stored in the $k$-th node of the
tree will be $R_k\remove S_k$, where $R_k$ and $S_k$ are the marginal
valuations defined over the
set of variables represented by the sets $h_k$ and $s_k$,
respectively. Each step is a kind of unidirectional message-passing
(towards the actual node $k$)
 in that 
a message is calculated at a node and then (1) removed from the valuation of
the node and (2) added to the node closer to $k$. 
The valuation of nodes $i=1,...,k$ at the beginning of step concerning node
$k$ is denoted with $V_{i,k}$. 
At the end of a step, the valuation is denoted with $V_{i,k-1}$ except for
node $k$ which is denoted with $R_k\remove S_k$\\

\noindent
{\bf The Algorithm: }

\noindent
{\bf begin}
\begin{enumerate}
\item for $k:=n$ step -1 downto 1  $V_{k,n}$:= $V_k$ 
\item for k:=n step -1 downto 2\\
   {\bf begin} 
   \begin{enumerate}
   \item Construct a subtree $\Gamma_k=(\bfH_k,\bfE_k)$ of T consisting only
of nodes 
$\bfH_k=\{1,...,k\}$. 
   \item Introduce the order $<_k$ compatible with the tree 
$\Gamma_k$, but such that the node $k$ is considered as its root
(the smallest element in $<_k$).
   \item Mark all nodes of the $\Gamma_k$ inactive 
   \item while the the direct successor of 
 node $k$ in ordering $<_k$  inactive\\
 if, in ordering $<_k$, all direct successors   of node $i$ are
active, then:\\
      begin
      \begin{enumerate}
      \item Active node $i$
      \item Denote all its direct successors   as inactive
      \item Let $j$ be direct  predecessor of $i$ in $<_k$
      \item Calculate 
$$
V'_{i,k} 
 := (V_{i,k} \otimes  (\otimes \{M_{l\rightarrow i}| 
 l \in  (N(i)_k-\{j\}\})  
$$

$$
 M_{i\rightarrow j}:= {V'_{i,k}}^{\downarrow j \cap i       }
$$
If $(i,k) \not\in \bfE_k$ calculate:
$$
 V_{i,k-1} :=  V'_{i,k}  \remove  M_{i\rightarrow j}
$$
where $N(i)_k$ stands for the set of neighbours of the node $i$ in the
Markov subtree $\Gamma_k$,
\end{enumerate}
end
\item Let $k+$ denote the direct successor of node $k$ in $<_k$\\
      Calculate 
$$
R_k 
 := V_{k,k} \otimes   M_{k+\rightarrow k}
$$

$$
 S_k := R_k ^{\downarrow k\cap k+ }
$$

$$
 V_{k+,k-1} :=  (V'_{k+,k}  \remove  M_{k+\rightarrow k})  \otimes   S_k 
$$
\end{enumerate}
{\bf end}
\item Calculate $R_1:=V_{1,1}$
\end{enumerate} 
{\bf end}

\begin{thm} \label{tm2}
If $R_i$ and $S_i$ have been calculated by the above algorithm for
the Markov tree T, then 
 $$R = \otimes \{V_i| i = 1..n\} = $$
$$=R_1 \otimes  (\otimes \{(R_i\remove S_i)| i
= 2..n\}) $$
where $R$ stands for the joint valuation defined over \uu{X}.          
\end{thm}
\AnfBeweis
In any subtree $\Gamma_k$ for any node $i$ with predecessor $j$ in $<_k$
except $k$ and $k+$ we have, due to Lemma   \ref{tm1}
$$                   
V_{i,k} \otimes 
  (\otimes \{V'_{l,k}| l \in  (N(i)_k-\{j\}\}) 
=$$ $$=V_{i,k} \otimes 
  (\otimes \{V_{l,k-1}\otimes M_{l \rightarrow i}| l \in  (N(i)_k-\{j\}\})
=$$ $$
=V'_{i,k} \otimes 
  (\otimes \{V_{l,k-1}| l \in  (N(i)_k-\{j\}\}) 
$$
hence update on  passage of activation does not change the joint valuation. \\
Also we have that
$$                 
V'_{k+,k} \otimes  V_{k,k} = 
(V'_{k+,k}  \remove  M_{k+\rightarrow k}) \otimes
 (V_{k,k} \otimes   M_{k+\rightarrow k})=$$
$$= (V'_{k+,k}  \remove  M_{k+\rightarrow k}) \otimes      R_k = $$
$$= ((V'_{k+,k}  \remove  M_{k+\rightarrow k})\otimes S_k) \otimes     (
R_k\remove S_k)=$$ $$= V_{k+,k-1} \otimes     ( R_k\remove S_k)
$$             
The theorem is then provable by induction (on k running from n to 1)

\EndBeweis

\begin{thm} \label{tm3}
In the previous theorem, 
  $R_i=R ^{\downarrow h_i}$
\end{thm}
\AnfBeweis
It is easily seen that 
$R_k$ is always the projection of the joint 
valuation of the subtree $\Gamma_k$ (compare the 
message passing algorithm of Shenoy and Shafer \shortcite{Shenoy:86}). 
Hence especially $R_n$ is the projection of $R$ onto node $n$.\\
Further, let 
$R_{\Gamma_k}= V'_{k,k} \times (\otimes \{V_{i,k} |i=1,2,...,k-1\} $
Then, due to \ax{CM1} we have:
$R_{\Gamma_{k-1}}= 
R_{\Gamma_k}^{1 \cup 2 \cup .... \cup k-1}$. 
This implies, by induction, that $R_k$ is the projection of $R$ 
onto node $k$ for every $k=1,2,...,n$. 

\EndBeweis

These two theorems \ref{tm2}, \ref{tm3} may be summarized as follows.

\begin{thm} \label{th31}
 Let T = (\bfH,\bfE) be a Markov
representative of 
 a VBS $(\uu{X}, \bfS,
 (\sigma_s)_{s\in \bfS},\otimes ,\downarrow )$.
Let $R_i$
stands for the valuation marginalized to the set $v_i$ of variables and $S_j$
stands for the marginal potential assigned to the separator of
the pair $\{h_i, h_j\}$. Then
 $$R = \otimes \{V_i| i = 1..n\} = 
R_1 \otimes  (\otimes \{(R_i\remove S_i)| i
= 2..n\}) $$
where $R$ stands for the joint valuation defined over \uu{X}.          
\end{thm}

Note that this theorem and the subsequent one 
are generalizations of  theorems presented by
Wierzcho\'{n}, \shortcite{Wierzchon:95}, in that the restricting condition 
that 
 the removal operation has to satisfy   the
property 
$(\rho \otimes \sigma)\remove 
(\delta\otimes \sigma) = (\rho\remove \delta)$ for any normal
valuations $\rho, \sigma$ and $\delta$
has been dropped. They represent also generalizations of properties of 
Dempster-Shafer belief functions presented in 
\cite{Klopotek:93f} and \cite{Klopotek:95i}.

 With the theorem \ref{th31} we can easily compute join valuations for
subsets being set theoretical union of members of the family \bfH.
This fact presents Theorem \ref{th32} below.

\begin{thm} \label{th32}
Let $\Gamma = (\bfN, \bfF)$ be a subtree of a
Markov tree T = (\bfH, \bfE) satisfying assumptions of Theorem 
\ref{th31}
Assume that for each node $h_j \in  \bfH$ the marginal valuation, $R_j$, has
been already computed. If $h_r$ stands for the root node in the
subtree $\Gamma$ then
$$
 R ^{\downarrow \cup \bfN} = 
\otimes \{V_i| i \in  \bfH\}^{\downarrow \cup \bfN} = $$
$$=R_r \otimes
(\otimes \{(R_i\remove S_i)| i \in  \bfN-\{v_r\}) \} $$
where $\cup \bfN$ stands the set theoretical union of all sets contained
in \bfN.                             
\end{thm}
\AnfBeweis
The result is straight-forward if we recall the axiom \ax{CM1}  and the
separator property of the Markov tree.
\EndBeweis

 Now, if $h \subseteq \cup \bfN$ then $R ^{\downarrow h}$ is computed as 
 $(R ^{\downarrow \cup \bfN})^{\downarrow h}$. Xu \shortcite{Xu:95}
proposed the local computation technique to find such a
marginal: it is simple consequence of Theorem \ref{th32} above and of Lemma
2.5 in \cite{Shenoy:94}.

\section{
Query processing in VBS
}

 The problem of query processing was formulated by Pearl \shortcite{Pearl:88}
first. In this approach we modify the original Bayesian belief
network by adding new nodes with appropriate edges. Consider for
instance the query $q = (x_1 \land x_2) \lor x_3$ - see (%
\cite{Pearl:88}, p.
224). Obviously, this $q$ introduces new subset $h = \{x_1, x_2, x_3\}$
to \bfH. In Pearl's approach we add two additional nodes joined to
the original network by three edges.

This approach suffers from several disadvantages. Adding new nodes 
to a belief network may change, even radically, the structure 
of the corresponding hypergraph which makes reconstruction of the Markov tree
necessary, which is time-consuming. In the process, also Markov tree may
change radically and practically all valuations have to be recalculated. 

\newstaff{
In practice, query $q$ may be frequently expressed
 in form of a single conjunction
of elementary (that is mutually exclusive) expressions or a disjunction of a
few conjunctions. E.g.:  the query $q = (x_1 \land x_2) \lor x_3$ 
may be restated as 
$q = (x_1 \land x_2 \land \lnot x_3) \lor x_3$ 
where conjuctions $(x_1 \land x_2 \land \lnot x_3)$ and $x_3$  
are mutually exclusive. 
It can be shown that calculation of such a  query  can be done without
modification of Markov tree. 
Under probabilistic settings as well as in DST, if $A$ and $B$ are 
mutually excluding conditions then $\rho(A\lor B)= \rho(A)+\rho(B)$. 
(In our example: 
$\rho((x_1 \land x_2 \land \lnot x_3) \lor x_3)=
\rho(x_1 \land x_2 \land \lnot x_3) + \rho(x_3)$. 
Therefore, 
}
in  our approach we must simply
  compute ${R}^{\downarrow h}$
(with $h$ being the set of variables appearing in the query $q$), and next
we should
find valuations over the set of configurations logically equivalent to $q$.
\newstaff{%
In our example we find valuation first for configuration: 
 $X_1=true,X_2=true, X_3=false,$ universe value for other variables, and
then for configuration $X_3=true,$  universe value for other variables.
}

 The only problem is to find the subtree $\Gamma$ 
 with
  $h \subseteq \cup \bfN$.
In
\cite{Wierzchon:95} it was shown that the minimal subtree, in the
sense that $\cup \bfN$ is as small as possible, can be found by applying
modified Graham's test. The modification concerns step (1) of
this test: a variable is deleted only if it does not belong to
the set $h$.

\newstaff{
Consider~e.g.~again~the~hypergraph~$H_1=$\linebreak
$\{\{X_1,X_7,X_8\}
, \{X_2,X_5,X_6,X_7\}
, \{X_3,X_6\} 
,$
$ \{X_4,X_5\}
,$\linebreak
$ \{X_8,X_9,X_{10}\}
, \{X_{10}, X_{11}, X_{12}\}
\}$  (see Fig.\ref{hypergraph}).
We see that variables $X_1,X_3,$
$X_4,X_{11},$
$X_{12}$ are contained in only one
edge, but $X_1,X_3$ are in $h$. We delete  only the other getting  
 the hypergraph 
$H_1'=
\{\{X_1,X_7,X_8\}
,$
$ \{X_2,X_5,X_6,X_7\}
,$
$ \{X_3,X_6\}
,$
$ \{X_5\}
,$
$ \{X_8,X_9,X_{10}\}
,$
$ \{X_{10}\}
\}$. But then hyperedge 
     $\{X_5\}$
is  contained in $\{X_2,X_5,X_6,X_7\}$, 
and  $\{X_{10}\}$ in  $\{X_8,X_9,X_{10}\}$. So we delete them getting
 $H_1''=
\{\{X_1,X_7,X_8\}
, \{X_2,X_5,X_6,X_7\}
, \{X_3,X_6\}$ $
, \{X_8,X_9,X_{10}\}
\}$. Now variables $X_2,X_5,X_9,X_{10}$ are contained in only one edge
each, however $X_2$ is in $h$. We get  $H_1'''=
\{\{X_1,X_7,X_8\}
,$
$ \{X_2,X_6,X_7\}
,$
$ \{X_3,X_6\}
,$
$ \{X_8\}
\}$. Now  hyperedge $\{X_1,X_7,X_8\}$ contains
 $\{X_8\}$, hence we get  $H_1''''=
\{\{X_1,X_7,X_8\}
,$
$ \{X_2,X_6,X_7\}
,$
$ \{X_3,X_6\}
\}$.
$X_8$ appears only in one edge, so we get finally $H_1'''''=
\{\{X_1,X_7\}
,$
$ \{X_2,X_6,X_7\}
,$
$ \{X_3,X_6\}
\}$.
 No further reduction by modified  Graham test is possible. 
We conclude that out of 6 hyperedges of $H_1$ only three
$ \{X_1,X_7,X_8\}
, \{X_2,X_5,X_6,X_7\}
, \{X_3,X_6\}$ are necessary for query answering calculations
 (see Fig.\ref{subhypergraph}), and that
$\{X_2,X_5,X_6,X_7\}$ may be projected to $\{X_2,X_6,X_7\}$,
and $\{X_1,X_7,X_8\}$ onto $\{X_1,X_7\}$. 
}

 This procedure is much more effective than that one
suggested by Xu \shortcite{Xu:95}, because this last method heavily depends
on the topology of a Markov tree.

We can, however, pose the question, whether or not the optimal subtree 
of the Markov tree T = (\bfH, \bfE) with hypertree \bfH~ covering  an
original hypergraph {\bfS }  would be "better" for query answering than 
 an optimal
hypertree cover T' = (\bfH', \bfE') of the
result of the above-mentioned modified Graham test run over the original 
 hypergraph $\bfS$. The
answer to this question is rather ambiguous. 
We can clearly construct examples where T' would be more
optimal than the subtree T in terms e.g. of the maximum number of nodes in an
edge. However, we must take into account that for each query
not only T' but also 
 the valuation for each node of the tree T' has to be calculated from
the entire hypergraph \bfS. But we do not need to do that
with subtrees of T, because we have to calculate the
$R_j$'s for a given tree T once and we do not need to recalculate them  when
selecting
a subtree, and if the subtree is small enough we save much calculation
compared with processing of T' (even if T' has a more optimal structure for
a given query).

 Concluding this paper we want to stress that this approach is
implemented in the VBS system designed by our group.


\begin{thebibliography}{99}
%
%
\bibitem[\protect\citeauthoryear{Bertele \&  Brioschi}{1972}]{Bertele:72%
}
U. Bertele and F. Brioschi. 
\newblock {\em Nonserial Dynamic Programming}. 
\newblock Academic Press, NY, 1972. 

\bibitem[\protect\citeauthoryear{Clifford \& Preston}{1961}]{Clifford:61%
}
A.H.~Clifford and \linebreak G.B.~Preston. 
\newblock The Algebraic Theory of Semigroups. 
\newblock {\em American Mathematical Society}, Providence, Rhode
Island, vol. 1,   (1961)

\bibitem[\protect\citeauthoryear{Cooper \& Herskovits}{1992}]%
{Cooper:92%
}
G.F. Cooper and E. Herskovits. 
\newblock A Bayesian method for
the induction of probabilistic networks from data. 
\newblock {\em Machine Learning}, 9:309-347, 1992.

\bibitem[\protect\citeauthoryear{Jensen et al.}{1990}]{Jensen:90%
}
F.V. Jensen, S.L. Lauritzen,  and K.G. Olesen.
\newblock 
Bayesian updating in causal probabilistic networks by local
computations. 
\newblock {\em Computational Statistics Quarterly}, 4: 269-282, 1990. 
\bibitem[\protect\citeauthoryear{Klopotek}{1994}]{Klopotek:93f}
M.A.~K{\l}opotek.
\newblock Beliefs in Markov Trees - From Local Computations to Local 
Valuation.
\newblock in: R. Trappl, ed.: {\em Proc. EMCSR'94} Vol.1. pages  351-358,
1994. 
\bibitem[\protect\citeauthoryear{Klopotek}{1995}]{Klopotek:95i}
 M.A.~K{\l}opotek.
\newblock 
On (Anti)Conditional Independence in Dempster-Shafer Theory
\newblock 
%
to appear in
{\em Journal Mathware and Softcomputing}, 1995.  




\bibitem[\protect\citeauthoryear{Lauritzen \& Spiegelhalter}{1988}]%
{Lauritzen:88%
}
S.L. Lauritzen and D.J. Spiegelhalter.
\newblock 
 Local computation with probabilities on graphical structures and
their application to expert systems. 
\newblock 
{\em J. Roy. Stat. Soc.}, B50:
pages 157-244, 1988.

\bibitem[\protect\citeauthoryear{Pearl}{1988}]{Pearl:88%
}
J. Pearl. 
\newblock {\em Probabilistic Reasoning in Intelligent
Systems: Networks of Plausible Inference. }
\newblock Morgan Kaufman, 1988.

\bibitem[\protect\citeauthoryear{Shafer}{1976}]{Shafer:76%
}
G. Shafer.
\newblock {\em A Mathematical Theory of Evidence}.
\newblock  Princeton
University Press, Princeton, NJ, 1976.

\bibitem[\protect\citeauthoryear{Shenoy}{1989}]{Shenoy:89%
}
P.P. Shenoy.
\newblock  A valuation-based language for expert
systems.
\newblock {\em  International Journal of Approximate Reasoning},
3:383-411, 1989.

\bibitem[\protect\citeauthoryear{Shenoy}{1991}]{Shenoy:91%
}
P.P. Shenoy.
\newblock  Valuation-based systems for discrete
optimization.
\newblock 
in P.P.
Bonissone%
, M. Henrion, L.N. Kanal and J.F. Lemmer,
 eds:
{\em Uncertainty in Artificial Intelligence}
6, 
North-Holland, Amsterdam, pages  385-400,  1991.

\bibitem[\protect\citeauthoryear{Shenoy}{1993}]{Shenoy:93%
}
P.P. Shenoy. 
\newblock
 A new method for representing and solving
Bayesian decision problems.
\newblock in D.J. Hand, ed.:
{\em Artificial Intelligence Frontiers in
Statistics: AI and Statistics III} , Chapman \&
Hall, London, pages 119-138,  1993.

\bibitem[\protect\citeauthoryear{Shenoy}{1994}]{Shenoy:94%
}
P.P. Shenoy. 
\newblock
 Conditional independence in valuation-based
systems. 
\newblock
{\em International Journal of Approximate Reasoning},
10:203-234, 1994.

\bibitem[\protect\citeauthoryear{Shenoy \& Shafer}{1986}]{Shenoy:86%
}
P.P. Shenoy and G. Shafer.  
\newblock
Propagating belief
functions using local computations.  
\newblock
{\em IEEE Expert}, 1(3), pages  43-52, 1986.

\bibitem[\protect\citeauthoryear{Thoma}{1991}]{Thoma:91%
}
H.M. Thoma.  
\newblock
Belief function computations.  
\newblock
in: I.R.
Goodman et al (Eds.): {\em Conditional Logics in Expert Systems},
North-Holland, pages  269-308, 1991.


\bibitem[\protect\citeauthoryear{Wen}{1991}]{Wen:91%
}
W.X. Wen.  
\newblock
From relational databases to belief networks, 
\newblock 
in: B.D'Ambrosio, Ph. Smets, and P.P. Bonissone (Eds.), {\em Proc.
7-th Conference on Uncertainty in Artificial Intelligence},
Morgan Kaufmann, pages  406-413, 1991.

\bibitem[\protect\citeauthoryear{Wierzchon}{1995}]{Wierzchon:95%
}
S.T. Wierzcho\'{n}. 
\newblock
 Markov-like properties of joint
valuations,  
\newblock
submitted, 1995.
 
\bibitem[\protect\citeauthoryear{Wong et al.}{1993}]{Wong:93%
}
S.K. Wong, Y. Xiang, and X. Nie. 
\newblock Representation of
Bayesian networks as relational databases.
\newblock  in: D. Heckerman, and
A. Mamdani, (Eds.), {\em Proc. 9-th Conference on Uncertainty in
Artificial Intelligence}, Morgan Kaufmann, pages  159-165, 1993. 

\bibitem[\protect\citeauthoryear{Xu}{1995}]{Xu:95%
}
H. Xu. 
\newblock Computing marginals for arbitrary subsets from
marginal representation in Markov trees. 
\newblock {\em Artificial
Intelligence}, 74, pages  177-189, 1995.

\end{thebibliography}
\end{document}